\documentclass[conference]{IEEEtran}
\IEEEoverridecommandlockouts

\usepackage{cite}
\usepackage{amsmath,amssymb,amsfonts}
\usepackage{algorithmic}
\usepackage{graphicx}
\usepackage{textcomp}
\usepackage{xcolor}
\usepackage{comment}
\usepackage{makecell}
\def\BibTeX{{\rm B\kern-.05em{\sc i\kern-.025em b}\kern-.08em
    T\kern-.1667em\lower.7ex\hbox{E}\kern-.125emX}}
\begin{document}

\title{Exploring the impact of Optimised Hyperparameters on Bi-LSTM-based Contextual Anomaly Detector\\
}

\author{\IEEEauthorblockN{1\textsuperscript{st} Aafan Ahmad Toor, 2\textsuperscript{nd} Jia-Chun Lin, 3\textsuperscript{rd} Ernst Gunnar Gran}
\IEEEauthorblockA{\textit{Department of Information Security and Communication Technology} \\
\textit{Norwegian University of Science and Technology}\\
Gjøvik, Norway \\
aafan.a.toor@ntnu.no, jia-chun.lin@ntnu.no, ernst.g.gran@ntnu.no}
}
\maketitle

\begin{abstract}
The exponential growth in the usage of Internet of Things in daily life has caused immense increase in the generation of time series data. Smart homes is one such domain where bulk of data is being generated and anomaly detection is one of the many challenges addressed by researchers in recent years. Contextual anomaly is a kind of anomaly that may show deviation from the normal pattern like point or sequence anomalies, but it also requires prior knowledge about the data domain and the actions that caused the deviation. Recent studies based on Recurrent Neural Networks (RNN) have demonstrated strong performance in anomaly detection. This study explores the impact of automatically tuned hyperparamteres on Unsupervised Online Contextual Anomaly Detection (UoCAD) approach by proposing UoCAD with Optimised Hyperparamnters (UoCAD-OH). UoCAD-OH conducts hyperparameter optimisation on Bi-LSTM model in an offline phase and uses the fine-tuned hyperparameters to detect anomalies during the online phase. The experiments involve evaluating the proposed framework on two smart home air quality datasets containing contextual anomalies. The evaluation metrics used are Precision, Recall, and F1 score.
\end{abstract}


\section{Introduction}\label{sec:intro}
With the advancement of sensor technology, more and more Internet of Things (IoT) devices are being used throughout the world and their numbers are growing exponentially \cite{fahrmann2024anomaly}. IoT devices are interconnected devices that are designed to collect and share data with each other and often with the cloud. A smart device is referred to as an IoT device which not only shares the data with the cloud but also improves the daily life. For instance, air quality sensors in a smart home not just records measures like temperature, CO2, PM2.5, humidity, etc., but also alerts the home owners of potential dangers like virus risk, air pollution, burglary, etc. Use of IoT devices is not limited to smart homes, many use cases of IoT can be found in industrial applications, healthcare, agriculture, factories, finance, bio-mechanics, etc. \cite{fahrmann2024anomaly, sgueglia2022systematic, mejri2024unsupervised}. The data generated by these IoT devices is also growing exponentially \cite{belay2023unsupervised}, which brings us to challenges of storing and processing this huge amounts of data, monitoring behaviors, and predicting patterns to improve the quality of life.

The data generated from IoT devices, referred to as time series, is time-bound and it is important to preserve its temporal context in order to extract useful information. Time series is a set of sequential data points taken as equally, or in some cases irregularly, spaced points in time. Time series can be univariate, i.e., consisting of only one time-bound column, or multivariate, i.e., multiple time-bound columns recording data for the same event. Time series often behaves unusually and show signs of irregularity due to system failure, sensor malfunction, or malicious activities. These irregularities are called anomalies which are essentially a deviation from the normal pattern. 

Researchers have identified different types of anomalies based on their nature and behavior, such as point anomalies (a significant deviation of a single data point), collective anomalies (a significant deviation of multiple adjacent data points), and contextual anomalies (a point or collective anomaly that is considered anomalous only within a specific context, such as the time of occurrence)~\cite{mejri2024unsupervised}. Some contextual anomalies do not deviate significantly from the normal pattern but their unusual temporal location makes them anomalous. These complex types of anomalies are difficult to detect \cite{toor2024uocad}. To better define such anomalies, defining the 'normal' in the context of a specific domain is also very important.  

Tuning the hyperparameters of a DNN-based model has been addressed in many recent studies \cite{nizam2022real, nguyen2024real, li2021improving, kim2023contextual, 10112268, abdallah2021hybrid, wang2024intelligent}. The objective of tuning the hyperparameters is to balance the bias-variance trade-off. Bias-variance trade-off ensures the adequate bifurcation of model predictions and the ground truth (bias), and the ability of the model to produce consistent and reliable results for unseen data (variance). For any DNN-based model, there are many hyperparameters which have direct or indirect affect on the prediction accuracy and overall performance. For example, learning rate affects the convergence of the model, having more neurons increases a model's ability to learn complex relationships among the data points, model's complexity and learning ability is determined by the number of layers, etc. 

This study builds on the work of the Unsupervised Online Contextual Anomaly Detector (UoCAD) \cite{toor2024uocad} by introducing UoCAD-OH, i.e., UoCAD with Optimised Hyperparameters. UoCAD is an online contextual anomaly detection method based on a Bidirectional LSTM (Bi-LSTM) that uses a sliding window approach to process smart home time series for anomaly detection. UoCAD conducts experiments using eight different window sizes with fixed hyperparameters for the Bi-LSTM model. However, like many state-of-the-art methods, UoCAD does not perform the hyperparameter tuning for the underlying model. UoCAD-OH addresses this by incorporating automatic hyperparameter tuning to improve the anomaly detection results.

The remainder of the paper is organized as follows. Relevant related works are presented in Section \ref{sec:related}. Details of the UoCAD and UoCAD-OH are provided in Section \ref{sec:uocad}. Section \ref{sec:experiments} discusses the dataset description, automatic hyperparameter optimisation details, experiments, and results, and finally, conclusions are drawn in Section \ref{sec:conc}.

\section{Related Works}\label{sec:related}
In recent years, Deep Neural Network (DNN) based methods have outperformed the conventional methods for time series anomaly detection \cite{LI202393, mejri2024unsupervised}. Within the DNN domain, Recurrent Neural Network (RNN) based methods, such as Simple RNN, LSTM, and GRU are most common among researchers for anomaly detection \cite{belay2023unsupervised}. LSTM and GRU are the improved forms of simple RNNs, and they have been used more frequently by researchers. However, simple RNNs have also been used \cite{hasnain2020performance} to detect performance anomalies from a simulated web service time series with throughput as the context of anomaly. A supervised learning-based study uses simple RNNs in combination with convolutional neural networks \cite{canizo2019multi} to detect anomalies in a time series representing the operating status of a service elevator. In this study, contextual anomalies specifically refer to the elevator's descending journeys. Both \cite{hasnain2020performance} and \cite{canizo2019multi} have compared simple RNN's performance with various RNN-based methods, including LSTM, GRU, Bi-LSTM, and Bi-GRU. 

LSTM-based anomaly detection approaches are very common among researchers \cite{pasini2022contextual, lee2021salad, lee2023repad2, lee2023rola}. The LSTMAD framework \cite{pasini2022contextual} consists of four modules: noise reduction, normalization, an LSTM layer, and anomaly detection. The framework captures the influence of context within the dataset and detects contextual anomalies. LSTMAD is evaluated on a synthetic dataset and a real world dataset of a smart card ticketing system. Fluctuations in the ticket sales caused by some event is considered as contextual anomalies in the real world dataset. Another LSTM-based method SALAD \cite{lee2021salad} detects anomalies from the recurrent time series in the New York City (NYC) taxi demand dataset. SALAD is an online and lightweight method that detects anomalies by calculating the Average Absolute Relative Error (AARE) and a self-adaptive detection threshold. Due to the resource requirements for calculating its detection threshold based on all the historical AARE values, SALAD becomes less efficient when dealing with open-ended time series. 

RePAD2 \cite{lee2023repad2} is a sliding-window-based real-time time series anomaly detection method that is less resource-intensive than SALAD. Instead of storing all historical AARE values, RePAD2 retains only the most recent values and determines the decision threshold based on these recent values. RePAD2 evaluated the performance on four different sliding window sizes and the results show that it outperformed it's predecessors in anomaly detection, retrain ratio, and time performance. A recent study \cite{lee2024impact} evaluated the performance of RePAD2 using three simple RNN variants—RNN, LSTM, and GRU—and three deep learning platforms: TensorFlow-Keras, PyTorch, and Deeplearning4j. The results indicated that the LSTM variant of RePAD2, implemented on Deeplearning4j, delivered the best performance.

RoLA \cite{lee2023rola} is another LSTM-based study that employs a divide-and-conquer strategy for anomaly detection from the FerryBox dataset, which was collected using sensors on a Norwegian tall ship. RoLA splits multivariate time series into univariate time series and introduces the concept of Lightweight Anomaly Detectors (LAD), which operate in a parallel processing environment to jointly detect anomalies in the multivariate time series.

One common characteristic of most of the studies discussed above is the absence of contextual anomalies. This highlights the gap in the availability of real-world datasets having contextual anomalies. We discuss some studies that focus on detecting contextual anomalies but their methodology is neither RNN nor DNN-based. These anomaly contexts include traffic increase on highway due to dodgers game \cite{hayes2015contextual}, stock market manipulation~\cite{golmohammadi2015time}, cyber attack on smart grid \cite{kosek2016contextual}, context inference using causal discovery~\cite{hela2018early}, contextual anomaly relationship on the basis of temporal and spacial neighbors in wireless sensor network data \cite{yu2020adaptive}, and calculation of the context using importance scores \cite{calikus2022wisdom}. 

The related works discussed above highlight several gaps in the field of online time series anomaly detection. First, only few studies focus on contextual anomalies, likely due to the scarcity of real-world datasets containing such anomalies. Second, very few studies perform automatic hyperparameter optimisation, instead most studies either tune only a few hyperparamteres or use predefined hyperparamteres based on hunch. Lastly, although most online anomaly detection approaches use a sliding window technique, only a small number recognize the importance of selecting an appropriate window size. Many studies rely on a single predefined window size, increasing the likelihood of selecting one that fails to accurately capture the underlying anomalies. This study addresses these issues and aims to fill these gaps.

\section{UoCAD and UoCAD-OH}\label{sec:uocad}
The Unsupervised Online Contextual Anomaly Detector (UoCAD)~\cite{toor2024uocad} was proposed to detect contextual anomalies in multivariate time series data from smart homes. The core idea behind UoCAD is to process time series data in online manner to identify contextual anomalies. UoCAD employs a sliding window approach to process small chunks of incoming data and determines whether the next instance is anomalous or normal. Each sliding window is preprocessed and fed into a Bi-LSTM model to compute error losses. 

UoCAD then calculates the Average Absolute Relative Error (AARE) values using Equation \ref{eq_aare}. Here, \textit{M} represents each feature, \textit{N} represent the sliding window size, $y_i$ and $\hat{y}_i$ denote the actual and predicted values, respectively. 
\begin{equation}\label{eq_aare}
AARE_M = \frac{1}{N} \sum_{i=1}^{N} | \frac{y_{i} - \hat{y}_{i}} {y_{i}} |
\end{equation}
Feature-wise dynamic thresholds are calculated by substituting the mean and standard deviation of the historical AARE values into Equation \ref{eq_thresh}. To detect real-world contextual anomalies, UoCAD utilizes smart home air quality data curated to include an actual contextual anomaly, namely 'unintended cooking.'
\begin{equation}\label{eq_thresh}
Thd_M = \mu_{AARE_{M}} + 3 \cdot \sigma_{AARE_{M}}
\end{equation}

UoCAD-OH adds the hyperparameter optimisation phase in the already established methodology of UoCAD. UoCAD-OH uses a hyperparameter tuner, which takes a relatively large multivariate time series dataset to perform automatic hyperparameter optimisation. The optimised hyperparameters for the Bi-LSTM model are then used by UoCAD-OH in the anomaly detection process. 

\section{Experiments and Results}\label{sec:experiments}
This section outlines the details of the experiments conducted in this study and presents the results obtained.

\subsection{Datasets}
The smart home time series datasets used in this study were collected using the AirThings View Plus device \cite{ViewPlus}. The data includes features related to indoor air quality in a smart home environment, with the device placed in the kitchen near the stove. The multivariate time series datasets consist of a timestamp feature and nine numeric features: temperature, humidity, carbon dioxide (CO2), volatile organic compounds (VOC), particulate matter (PM) 2.5 and 1.0, pressure, light, and sound. Three datasets from different periods were used for the experiments: a 2-day dataset containing one contextual anomaly, a 5-day dataset with two contextual anomalies, and a 5-month dataset. 
Further details on these datasets are provided in the following subsections.

\subsubsection{2-days dataset (2d1a)}
This multivariate time series dataset, referred to as 2d1a, is also used in UoCAD \cite{toor2024uocad}. It is a 2-day air quality dataset containing 1,151 instances. The dataset includes one sequential contextual anomaly—unintended cooking—which spans 28 instances. This anomaly was generated by intentionally burning food on the stove while ventilation was turned off, and the doors and windows were shut. Table \ref{tab:2d1a_specs} \cite{toor2024uocad} provides a summary of the dataset, including the minimum, maximum, average, and standard deviation values for each feature.

\begin{table}[t]
\footnotesize
\centering
\caption{Summary of the 2d1a dataset \cite{toor2024uocad}.}\label{tab:2d1a_specs} 
\begin{tabular}{|c|c|c|c|c|}
  \hline
  \textbf{Feature} & \textbf{Min.} & \textbf{Max.} & \textbf{Avg.} & \textbf{Std. Dev.} \\
  \hline
  Temp & 18.79 & 27.89 & 22.0 & 1.97 \\
  \hline
  Humidity & 44.62 & 95.76 & 58.53 & 9.51 \\
  \hline
  Pressure & 968.0 & 981.0 & 974.5 & 4.32 \\
  \hline
  CO2 & 635 & 2518 & 1757.5 & 434.54 \\
  \hline
  VOC & 46 & 721 & 188.30 & 121.65 \\
  \hline
  Light & 0 & 74 & 21.39 & 25.01 \\
  \hline
  PM10 & 1 & 659 & 97.95 & 100.81 \\
  \hline
  PM2.5 & 1 & 852 & 101.19 & 106.32 \\
  \hline
  Sound & 37 & 94 & 50.80 & 12.13 \\
  \hline
\end{tabular}
\end{table}

\subsubsection{10-days dataset (10d2a)}

The second dataset used in this study was collected from the same AirThings View Plus device. Referred to here as 10d2a, this dataset consists of 10 days of air quality sensor data and includes two sequential contextual anomalies. The first anomaly simulates a heating malfunction. To generate this anomaly, the heating was turned off, and the doors and windows were opened while the outside temperature ranged between -5°C and -8°C. The second anomaly, occurring two days after the first, is unintended cooking, generated in the same manner as in the 2d1a dataset. The dataset comprises 6,336 instances, with each anomaly spanning 24 instances. Table \ref{tab:10d2a_specs} provides a summary of the dataset.

\begin{table}[t]
\footnotesize
\centering
\caption{Summary of the 10d2a dataset.}\label{tab:10d2a_specs}
\begin{tabular}{|c|c|c|c|c|}
  \hline
  \textbf{Feature} & \textbf{Min.} & \textbf{Max.} & \textbf{Avg.} & \textbf{Std. Dev.} \\
  \hline
  Temp & 8.99 & 33.11 & 21.75 & 2.83 \\
  \hline
  Humidity & 23.22 & 95.79 & 51.36 & 9,50 \\
  \hline
  Pressure & 983 & 1010 & 993.48 & 6.2 \\
  \hline
  CO2 & 400 & 2402 & 1439.9 & 357.61 \\
  \hline
  VOC & 0 & 4851 & 133.17 & 290.83 \\
  \hline
  Light & 0 & 44 & 11.68 & 13.24 \\
  \hline
  PM1.0 & 0 & 697 & 83.06 & 116.71 \\
  \hline
  PM2.5 & 0 & 1000 & 97.48 & 159.71 \\
  \hline
  Sound & 37 & 84 & 50.76 & 11.93 \\
  \hline
\end{tabular}
\end{table}

\subsubsection{5-months dataset (5M)}

The third multivariate time series dataset used in this study, referred to as 5M, is a large dataset containing 91,738 instances. This dataset is used only for training the models for automatic hyperparameter optimisation. It does not contain labeled anomalies, as it is not intended for anomaly detection but specifically for hyperparameter tuning. The data was collected over a five-month period, which does not overlap with the 2d1a and the 10d2a datasets. The AirThings View Plus device was located in the same place as for the other two datasets. A summary of this dataset is provided in Table~\ref{tab:5m_specs}.

\begin{table}[t]
\footnotesize
\centering
\caption{Summary of the 5M dataset.}\label{tab:5m_specs}
\begin{tabular}{|c|c|c|c|c|}
  \hline
  \textbf{Feature} & \textbf{Min.} & \textbf{Max.} & \textbf{Avg.} & \textbf{Std. Dev.} \\
  \hline
  Temp & 0 & 58.2 & 24.49 & 2.86 \\
  \hline
  Humidity & 0 & 122.08 & 42.86 & 9.01 \\
  \hline
  Pressure & 946.8 & 2008 & 991.61 & 11.28 \\
  \hline
  CO2 & 0 & 2679 & 1010.92 & 417.37 \\
  \hline
  VOC & 46 & 4851 & 198.46 & 207.81 \\
  \hline
  Light & 0 & 76 & 17.05 & 15.77 \\
  \hline
  PM1.0 & 0 & 706 & 42.89 & 84.08 \\
  \hline
  PM2.5 & 0 & 1000 & 48.25 & 106.88 \\
  \hline
  Sound & 0 & 133 & 53.18 & 11.34 \\
  \hline
\end{tabular}
\end{table}

\subsection{Hyperparameter Optimisation}
Hyperparameter tuning is an offline task that exhaustively test a lot of parameters and their respective values to find an optimised combination of hyperparameters. Hyperparameter tuning is model- and dataset-specific, so in this study, the tuning process was repeated for Bi-LSTM model using the 5M dataset. The Keras Tuner \cite{omalley2019kerastuner} was employed to automatically tune the hyperparameters for each model. Table \ref{tab:tuner_setup} provides the details of the tuner setup.

\begin{table}[t]
\footnotesize
\centering
\caption{Hyperparameter tuning setup}\label{tab:tuner_setup}
\begin{tabular}{| c | c |}
  \hline
  Method & Hyperband  \\ 
  \hline
  Batch Size & 100  \\
  \hline
  Epochs & 50  \\
  \hline
  Max Trials & 1 \\
  \hline
  Execution per Trial & 1 \\
  \hline
  Max retries per trial & 2 \\
  \hline
  Objective & Minimum loss \\
  \hline
\end{tabular}
\end{table}

The hyperband search method is used for the tuning process. This method was chosen for its efficiency, as opposed to its alternative method, i.e., Random Search. Random Search method trains the model with all possible combinations of hyperparameters and their values, which is time-consuming and often results in trying many suboptimal combinations. In contrast, Hyperband begins with a random set of hyperparameters and then intelligently selects the best ones after running a few epochs. This reduces the search space, allowing the tuner to iteratively find a near-optimal set of hyperparameters more efficiently.

\begin{table}[t]
\footnotesize
\centering
\caption{Hyperparameter choices/values}\label{tab:hp_choices}
\begin{tabular}{| c | c |}
  \hline
  Input units & min\_value=32, max\_value=192, step=32  \\ 
  \hline
  Activation function & ReLU, LeakyReLU, sigmoid, softmax  \\
  \hline
  Learning rate & 1e-2, 1e-3, 1e-4  \\
  \hline
  Optimiser function & rmsprop, adam, adadelta, adagrad \\
  \hline
  Number of layers & min=2, max=5 \\
  \hline
  Dropout rate & min=0.1, max=0.6, step=0.1 \\
  \hline
\end{tabular}
\end{table}

Batch size and epoch count were set to 100 and 50 respectively, with early stopping configured to monitor validation loss. These are commonly used settings for a medium level dataset, such as 5M used in this study. The "Max Trials" and "Execution per Trial" options were set to 1, which means that tuner builds and executes model only once with each combination of parameters. Due to the deterministic nature of deep learning models, running the same data and model architecture can yield slightly different results each time. For this reason, running multiple trials can sometimes provide more optimal results. However, executing multiple trials also increased time and memory requirements. The "Max retires per Trial" option enables tuner to retry the trial in case it crashes or produces invalid results. Finally, the "Objective" option was set to monitor minimum loss, which provides the tuner to evaluate the model's performance.

\begin{table}[t]
\footnotesize
\centering
\caption{Optimised hyperparameters for Bi-LSTM model}\label{tab:tuning_results}
\begin{tabular}{| m{14.0em} | m{8.0em} |}
  \hline
  Hyperparameters & Optimised Values \\ 
  \hline
  Input units & 160 \\
  \hline
  Activation  & ReLU  \\
  \hline
  Learning rate & 0.0001 \\
  \hline
  Optimiser  & Adam \\
  \hline
  Number of layers & 2 \\
  \hline
  Dropout rate & 0.2 \\
  \hline
  Epoch count  & 50 \\
  \Xhline{2\arrayrulewidth}
  Execution time in minutes & 23.94 \\
   \hline
\end{tabular}
\end{table}

Attaining best results from a deep learning model requires choosing an appropriate number of input units and layers, a combination of activation function, learning rate, and optimiser function, and an adequate regularization mechanism such as dropout layer. Hyperparameter choices and their respective values are presented in Table \ref{tab:hp_choices}. These are not only the most commonly optimisable hyperparameter choices, but also these choices are most important for tuning a deep learning model. Results of the hyperparameter tuning are presented in Table \ref{tab:tuning_results}. These results are produced by running the tuner on 5M dataset, with hyperparameter choices and values for each of the deep learning models selected for this study. Optimised hyperparameters obtained from these experiments are later used to detect anomalies from the 2d1a and the 10d2a datasets.

\subsection{Evaluation Metrics}
The performance of the Bi-LSTM model for anomaly detection is evaluated using the Precision: $\frac{TP}{TP + FP}$, Recall: $\frac{TP}{TP + FN}$, and F1-score: $2 \cdot \frac{Precision \cdot Recall}{Precision + Recall}$ metrics, which most researchers prefer for anomaly detection. Here, TP refers to True Positives, FP to False Positives, and FN to False Negatives. Sequence anomalies are a collection of adjacent instances representing a particular event; thus, detecting any one of these adjacent instances indicates that the entire anomalous event has been detected. Based on this assumption, if $p$ is the total number of anomalous instances and any instance $i$ belongs to $p$, the whole $p$ is considered as  true positive. 

\subsection{Experimental Setup}
All experiments are conducted on an Apple MacBook Pro with an M2 chip, 16GB of RAM, and 256GB of disk storage. The code was written using Python-based libraries, including TensorFlow \cite{15abadi2016tensorflow} for end-to-end machine learning and Keras \cite{16chollet2015keras} for neural network methods.

To ensure consistency with UoCAD's experiments and allow for a fair evaluation, this study conducts experiments using eight window sizes for each selected model: 6, 12, 24, 48, 72, 96, 120, and 144. Similarly, this study also presents anomaly detection results using two criteria: individual and majority. Under the individual criterion, an anomaly is detected if at least one feature reports an anomaly, whereas, in the majority criterion, an anomaly is considered valid only if at least five out of nine features report it.

\begin{table}[htbp]
\scriptsize
\caption{Detection performance of UoCAD-OH in different scenarios. 
}\label{tab:uocad_blstm} \centering
\begin{tabular}{|c|c|c|c|c|c|c|c|c|c|c|}
  \hline
   & \multicolumn{3}{|c|}{2d1a} & \multicolumn{6}{c|}{10d2a}  \\
   \cline{5-10}
   & \multicolumn{3}{|c|}{} & \multicolumn{3}{|c|}{Anomaly 1} & \multicolumn{3}{|c|}{Anomaly 2} \\
  \cline{2-10}
  \textbf{Combs.} & \textbf{P} & \textbf{R} & \textbf{F1} & \textbf{P} & \textbf{R} & \textbf{F1} & \textbf{P} & \textbf{R} & \textbf{F1} \\
   \hline
  Ind-6  & 0.89 & 1.0 & 0.94 & 0 & 0 & 0 & 0.26 & 1.0 & 0.41 \\
  \hline
  Ind-12  & 0.93 & 1.0 & 0.96 & 0 & 0 & 0 & 0.24 & 1.0 & 0.39  \\
  \hline
  Ind-24  & 0.95 & 1.0 & 0.97 & 0.46 & 1.0 & 0.63 & 0.38 & 1.0 & 0.55 \\
  \hline
  Ind-48  & 0.81 & 1.0 & 0.89 & 0.18 & 1.0 & 0.30 & 0.17 & 1.0 & 0.29 \\
  \hline
  Ind-72  & 0.42 & 1.0 & 0.59 & 0.27 & 1.0 & 0.42 & 0 & 0 & 0 \\
  \hline
  Ind-96  & 0.44 & 1.0 & 0.61 & 0.13 & 1.0 & 0.23 & 0 & 0 & 0 \\
  \hline
  Ind-120  & 0 & 0 & 0 & 0.13 & 1.0 & 0.27 & 0 & 0 & 0 \\
  \hline
  Ind-144  & 0.81 & 1.0 & 0.89 & 0.21 & 1.0 & 0.35 & 0 & 0 & 0 \\
  \hline
  Maj-6  & 0 & 0 & 0 & 0 & 0 & 0 & 1.0 & 1.0 & 1.0 \\
  \hline
  Maj-12  & 0 & 0 & 0 & 0 & 0 & 0 & 0 & 0 & 0 \\
  \hline
  Maj-24  & 0 & 0 & 0 & 0.84 & 1.0 & 0.91 & 0 & 0 & 0 \\
  \hline
  Maj-48  & 0 & 0 & 0 & 0 & 0 & 0 & 0 & 0 & 0 \\
  \hline 
  Maj-72  & 0 & 0 & 0 & 0 & 0 & 0 & 0 & 0 & 0 \\
  \hline 
  Maj-96  & 0 & 0 & 0 & 0.57 & 1.0 & 0.72 & 0 & 0 & 0  \\
  \hline
  Maj-120 & 0 & 0 & 0 & 0.64 & 1.0 & 0.78 & 0 & 0 & 0 \\
  \hline
  Maj-144 & 0 & 0 & 0 & 0 & 0 & 0 & 0 & 0 & 0 \\
  \hline
\end{tabular}
\end{table}

\begin{figure}[htbp]
\centering
\includegraphics[width=\linewidth]{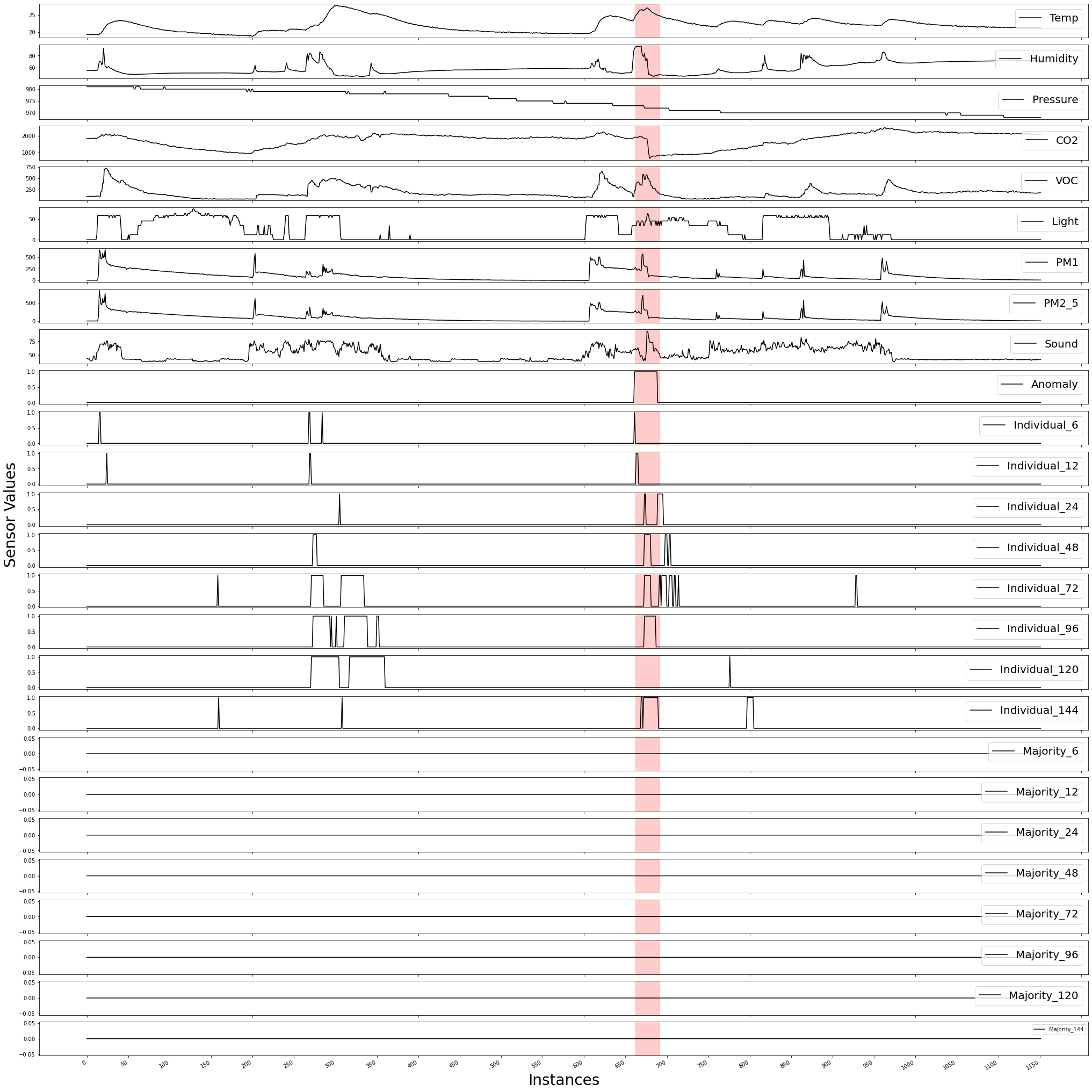}
\caption{Anomaly detection results of UoCAD-OH for different window sizes from the 2d1a dataset.}
\label{fig:uocad-blstm-2d}
\end{figure}

\begin{figure}[htbp]
\centering
\includegraphics[width=\linewidth]{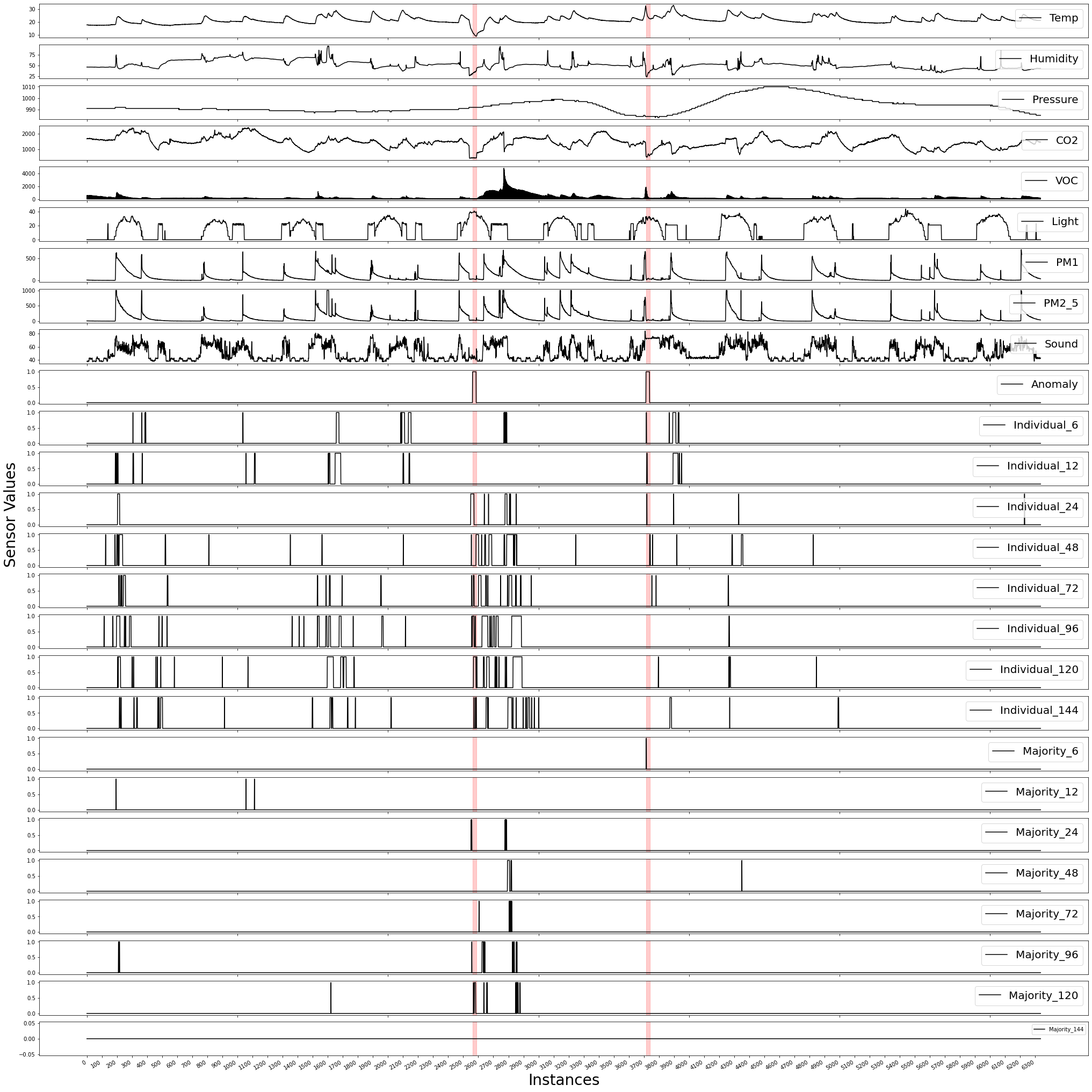}
\caption{Anomaly detection results of UoCAD-OH for different window sizes from the 10d2a dataset.}
\label{fig:uocad-blstm-10d}
\end{figure}

\subsection{Results and Discussion}
In this section, we present the anomaly detection results and discuss the performance of UoCAD-OH for the 2d1a and the 10d2a datasets. Table \ref{tab:uocad_blstm} summarize the Precision (referred as P), Recall (referred as R), and F1-scores (referred as F1) from the experiments, while Figures \ref{fig:uocad-blstm-2d} to \ref{fig:uocad-blstm-10d} visualize the anomaly detection results of UoCAD-OH for both datasets. 

It is important to note that UoCAD-OH has performed well in anomaly detection like its predecessor, i.e., UoCAD, which is also based on Bi-LSTM model. According to the results reported in the Table \ref{tab:uocad_blstm}, for the 2d1a dataset's Ind-6, Ind-12, Ind-24, Ind-48, Ind-72, Individual-96, and Ind-144 combinations, UoCAD-OH was able to detect anomalies with best Precision of 0.95, Recall of 1.0, and F1-score of 0.97. However, no majority combination could detect anomaly. For the 10d2a dataset, Ind-24 and Ind-48 combinations were able to detect both anomalies. Similarly, combinations like Ind-6, Ind-12, Ind-72, Ind-96, Ind-120, Maj-6, Maj-24, Maj-96, and Maj-120 reported detection of either Anomaly 1 or 2.

Ind-120 and Ind-144 are the only combinations for the 2d1a dataset that were least successful in detecting anomaly. Hence, we can deduce that the upper limit of window size for the 2d1a dataset is 96 for UoCAD=OH. In case of majority criteria, the combinations Maj-96, Maj-120, and Maj-144 proved to be worst as UoCAD-OH could not detect anomalies from the 2d1a dataset using these combinations. However, UoCAD-OH was more successful in detecting Anomaly 1 or 2 from the 10d2a dataset for selective window sizes.

It is important to note that the anomaly in the 2d1a dataset, "unintended cooking," is the same type of anomaly as Anomaly 2 in the 10d2a dataset. However, Anomaly 1 in the 10d2a dataset, "heating malfunction," differs in nature from "unintended cooking." Specifically, "unintended cooking" leads to a drop in CO2 and an increase in temperature, humidity, PM1.0, PM2.5, and VOC, whereas "heating malfunction" causes a rise in CO2 and a decrease in temperature, humidity, PM1.0, PM2.5, VOC, and pressure. Given these differences, an anomaly-wise comparison of detection results shows that UoCAD-OH produced similar results for the "unintended cooking" anomaly across both datasets. Notably, UoCAD-OH, performed better using the majority criterion for detecting the "heating malfunction" anomaly than for "unintended cooking."
\section{Conclusions}\label{sec:conc}
This study evaluates the anomaly detection performance of Recurrent Neural Network (RNN)-based Bi-LSTM model on smart home time series data containing contextual anomalies. UoCAD-OH extends the original UoCAD method by introducing both offline and online phases. In the offline phase, the hyperparameters of Bi-LSTM model are fine-tuned using a large smart home time series dataset. In the online phase, which is based on the original UoCAD method, the predefined hyperparameters are replaced with automatically fine-tuned hyperparameters. This study also compares the performance of eight sliding window sizes with UoCAD-OH, using both individual (where an anomaly is detected if at least one feature reports it) and majority (where more than half of the features must report an anomaly) criteria. 

The results show that the majority criterion is generally ineffective for anomaly detection. UoCAD-OH is capable of detecting anomalies for window sizes between 6 and 96, with 24 and 48 being the most effective window sizes. This study also provides a better understanding of the proposed method's ability to detect different types of contextual anomalies. In future, an expanded study can be performed to evaluate different RNN-based methods to find the best RNN variant. Additionally, UoCAD-OH can be made adaptive to different types of anomalies and scalable across various time series domains to enhance its generalization and robustness.

\bibliographystyle{IEEEtran}  
\bibliography{UoCAD-OH_arxiv_2025}  

\end{document}